\documentclass[10pt,twocolumn,letterpaper]{article}

\usepackage{cvpr}              

\usepackage{multirow}
\usepackage{booktabs}
\usepackage{MnSymbol} 
\usepackage[table,svgnames]{xcolor}
\usepackage{algorithm}
\usepackage{algpseudocode}
\usepackage{subcaption}
\usepackage{amssymb}
\usepackage{ulem}
\usepackage{tabularx}
\usepackage{tikz}
\usepackage[accsupp]{axessibility}  

\usetikzlibrary{arrows.meta,positioning,shapes.geometric,fit}

\definecolor{gray}{HTML}{d6d6ff}
\definecolor{lightgray}{HTML}{eeeeff}
\newcolumntype{M}[1]{>{\centering\arraybackslash}m{#1}}
\newcolumntype{L}[1]{>{\arraybackslash}m{#1}}

\definecolor{cvprblue}{rgb}{0.21,0.49,0.74}
\usepackage[pagebackref,breaklinks,colorlinks,allcolors=cvprblue]{hyperref}

\def\paperID{} 
\def\confName{CVPR}
\def\confYear{2025}

\title{Bézier Degradation Modeling for LiDAR-based Human Motion Capture}

\author{
    Xiaoqi An\textsuperscript{1} \quad
    Lin Zhao\textsuperscript{1}\thanks{Corresponding authors.} \quad
    Jun Li\textsuperscript{1} \quad
    Chen Gong\textsuperscript{1} \quad 
    Jian Yang\textsuperscript{1,2}\footnotemark[1]\\
\textsuperscript{1}PCA Lab, School of Computer Science and Engineering, Nanjing University of Science and Technology\\
\textsuperscript{2}PCA Lab, School of Intelligence Science and Technology, Nanjing University\\
{\tt\small \{xiaoqi.an,linzhao,junli,chen.gong,csjyang\}@njust.edu.cn}
}

\begin{document}
\maketitle
\begin{abstract}
LiDAR-based 3D human motion capture has broad applications in fields such as autonomous driving and robotics, where accurate motion reconstruction is crucial. However, existing methods often struggle with unstable inputs and severe occlusions, leading to jittery or even failed pose predictions. To address these challenges, we propose \textbf{\mbox{BMLiCap}}, a coarse-to-fine framework that models motion using temporally compressible Bézier curves. By reducing control points through a trajectory-preserving strategy, we obtain a coherent and learning-friendly motion representation. To reconstruct human actions from LiDAR point-cloud cues, we design a progressive motion-reconstruction module. Specifically, a Time-scale Motion Transformer (TMT) is introduced to predict motion curves at multiple temporal scales, and a Multi-level Motion Aggregator (MMA) is utilized to adaptively fuse the multi-scale curves to recover detailed, temporally coherent poses, effectively bridging observation gaps caused by occlusions and noise. Across four mainstream benchmarks LiDARHuman26M, FreeMotion, NoiseMotion, and SLOPER4D, BMLiCap achieves state-of-the-art accuracy and temporal continuity in complex scenes, demonstrating its ability to compensate for severe occlusions and reduce prediction jitter.
\end{abstract}    
\section{Introduction}
\label{sec:intro}

\begin{figure}[t]
  \centering
  \includegraphics[width=1.0\linewidth]{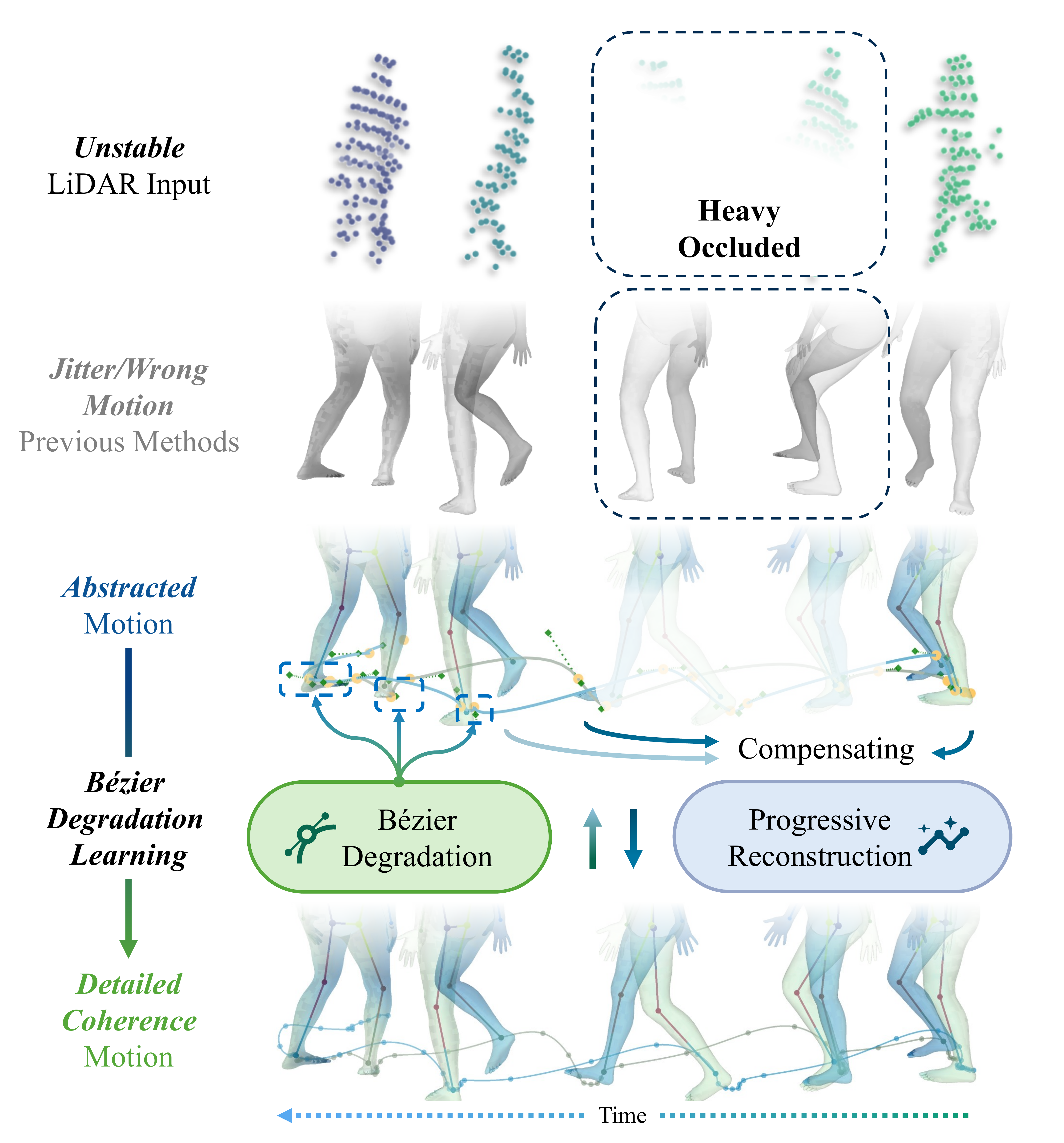}
   \caption{BMLiCap models human motion by Bézier curves. Our contributions mainly include: \textbf{(a)} a novel Bézier degradation method for generating easy-to-learn motion representations; \textbf{(b)} a progressive motion reconstruction model conditioned on LiDAR point clouds in a coarse-to-fine manner, compensating for severe input occlusions.  }
   \label{fig:intro}
   \vspace{-1em}
\end{figure}

3D human motion capture is a fundamental task in computer vision, which aims to reconstruct normalized 3D representations of the human body over time from sensor data. This technology is widely used in human-centric downstream applications, such as autonomous driving \cite{jeongMultiagentLongterm3D2024,landryPredictingPedestrianCrossing2025,ghiyaSGNetPoseStepwiseGoalDriven2025,lianCrossModalDrivenObject2025}, augmented reality \cite{xiaEnvPoserEnvironmentawareRealistic2025a,tangUnifiedDiffusionFramework2024}, and robotics \cite{vendrowJRDBPoseLargeScaleDataset2023}.

Traditional motion capture approaches often rely on wearable devices, where human poses are obtained via marker systems or a set of IMU sensors \cite{huangDeepInertialPoser2018}. With the development of computer vision techniques, low-cost optical alternatives based on RGB or RGB-D inputs \cite{bogoKeepItSMPL2016,liHybrIKXHybridAnalyticalNeural2025,zhuPseudoViewRepresentation2022,pesaventoANIMAccurateNeural2024,chanTransformerbasedAdaptivePrototype2024} have been proposed. Although these approaches achieve great success, their performance is typically restricted to indoor environments or lacks absolute depth. In recent years, the growing demands of autonomous driving and robotics have motivated research on human motion understanding in large-scale, unconstrained scenes. LiDAR-based motion capture has become a promising research direction \cite{renLiDARaidInertialPoser2023,fanLiDARHMR3DHuman2025,anPretrainingDensityAwarePose2025}, owing to its robustness to lighting conditions and reliable global depth.

However, LiDAR sensors inherently capture only sparse depth information from a monocular view of the human body, making it highly susceptible to occlusions and noisy point clouds. To address these issues, LiveHPS \cite{renLiveHPSLiDARbasedScenelevel2024} leverages features derived from SMPL \cite{loperSMPLSkinnedMultiperson2015} vertices as teacher signals to handle partial point cloud observations. Besides, LiveHPS++ \cite{renLiveHPSRobustCoherent2024} introduces velocity prediction to suppress noisy measurements. While these methods have made progress in learning human point cloud priors, as shown in  Fig.\ref{fig:intro}, they still struggle with long-term occlusions of critical joints and often produce jittery, biased predictions.

\begin{figure}[t]
    \centering
    \begin{subfigure}[b]{0.70\linewidth}
        \centering
        \includegraphics[width=\linewidth]{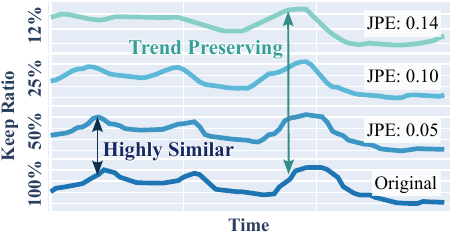}
        \caption{Example trajectory degradation}
        \label{fig:rmse_traject}
    \end{subfigure}
    \begin{subfigure}[b]{0.28\linewidth}
        \centering
        \includegraphics[width=\linewidth]{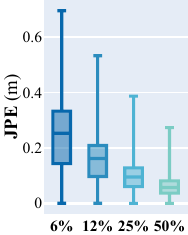}
        \vspace{-1em}
        \caption{Errors on dataset}
        \label{fig:rmse_boxplot}
    \end{subfigure}

    \caption{Analysis of motion approximation error using Bézier curves with different ratios of control points, indicating its robustness against occlusion.}
    \label{fig:rmse_analysis}
    \vspace{-1em}
\end{figure}

To tackle these challenges, we take a kinematics-driven approach. Rather than learning directly from incomplete point cloud features, we model human motion using Bézier curves. This parameterization explicitly exposes position, velocity, and acceleration, yielding smooth and stable interpolation even under long-term occlusions. As shown in Fig.\ref{fig:rmse_traject} and \ref{fig:rmse_boxplot}, aggressively pruning control points still preserves the global motion trend. This principle aligns with natural human movement. For instance, as illustrated in Fig.\ref{fig:intro}, a sequence of leg-lifting, stepping-forward, landing, and pushing-off actions can be coarsely summarized as “walking from A to B”. Hence, Bézier curves offer a hierarchical representation in which coarse trends capture intent and additional control points refine details. 

To reconstruct coherent human motion from unstable LiDAR observations, each stage of the Bézier hierarchical representation needs to be utilized. We therefore propose a coarse-to-fine reconstruction strategy along the temporal dimension, which is an inverse process of removing Bézier control points. At coarse stages, we first generate point-to-point motion trends of each joint, and then progressively refine them into finer sub-motions until detailed pose sequences at every timestep are obtained. This design not only exploits available visual cues from LiDAR features but also allows coarse-level motion trends to bridge the “observation breaks” caused by occlusions.

Overall, we introduce BMLiCap, a robust \uline{\textbf{Li}}DAR-based 3D motion \uline{\textbf{Cap}}ture framework with \uline{\textbf{B}}ézier \uline{\textbf{M}}otion representation. Specifically, we first propose a Bézier-curve-based temporal degradation scheme that progressively reduces control points, producing multi-level motion representations for training. Then, we design a Time-scale Motion Transformer (TMT) to predict motion curves at different temporal scales conditioned on LiDAR inputs. Finally, a Multi-level Motion Aggregator (MMA) is utilized to adaptively fuse motion representations into a complete fine-grained motion sequence. Our approach effectively mitigates jittering artifacts, enabling continuous motion perception. Extensive experiments demonstrate that BMLiCap significantly improves both motion accuracy and temporal consistency.

In summary, our contributions are threefold:

\begin{itemize}

    \item We propose representing human motion with Bézier curves and introducing a hierarchical degradation strategy. This strategy generates interpretable and learning-friendly multi-level motion representations.

    \item We introduce a progressive motion reconstruction approach to model the information flow between LiDAR features and multi-level motion representations, which reconstructs human motion in a coarse-to-fine way.

    \item We validate our method on four LiDAR-based motion benchmarks (LiDARH26M, FreeMotion, NoiseMotion, SLOPER4D), our BMLiCap achieves state-of-the-art performance in complex scenes with notable improvements in both accuracy and temporal continuity.
\end{itemize}
\section{Related works}
\label{sec:related_works}

\subsection{LiDAR-Based Human Motion Capture}

Motion capture is first introduced by early marker-based systems \cite{vlasic2007practical,vicon2010mocap}, then evolve to markerless approaches \cite{liu2013markerless,sridh2015MOT,rhodinEgoCapEgocentricMarkerless2016,xuFlyCapMarkerlessMotion2018,kankoConcurrentAssessmentGait2021}. Despite the high-accuracy reconstruction, they request expensive equipment and complex calibration. To alleviate these issues, monocular methods have been proposed, mainly devided into optimization based \cite{bogoKeepItSMPL2016,gulerDensePoseDenseHuman2018,kanazawa2019learning,gulerHoloPoseHolistic3D2019,liNIKINeuralInverse2023,liHybrIKXHybridAnalyticalNeural2025} or regression based \cite{varolBodyNetVolumetricInference2018,kanazawaEndtoendRecoveryHuman2018,liHybrIKHybridAnalyticalNeural2020,Zeng_2020_CVPR,zhangLearning3DHuman2022,linEndtoEndHumanPose2021,goelHumans4DReconstructing2023,Kim_2023_CVPR,dwivediTokenHMRAdvancingHuman2024,liHybrIKXHybridAnalyticalNeural2025} approaches. However, they remain limited by light sensitivity and depth ambiguity. Depth-camera solutions \cite{bashirovRealTimeRGBDBasedExtended2021,zhuPseudoViewRepresentation2022,pesaventoANIMAccurateNeural2024} partially address this but are unsuitable for outdoor use. In contrast, inertial methods \cite{huangDeepInertialPoser2018,patilFusionMultipleLidars2020,yiTransPoseRealtime3D2021,renLiDARaidInertialPoser2023,chenMotionCaptureInertial2024} avoid lighting constraints and occlusions by employing multiple IMUs. Yet, they still suffer from drift \cite{valtonenornhagTrustYourIMU2022} and rely on wearable devices.

With the ability of accurately capturing global depth in open environments, LiDAR-based motion capture has recently being exlored \cite{zhangLiDARCapV23DHuman2024,nisarPSASSLPoseSizeaware2025}. Some previous works were performed on dense point clouds \cite{moonV2VPoseNetVoxeltoVoxelPrediction2018,liuVoteHMROcclusionAwareVoting2021,caiPointHPSCascaded3D2023,zhangCoMPREfficientPoint2026}, but they have difficulty adapting to point clouds the LiDAR sensor provided. LiDARCap \cite{liLiDARCapLongrangeMarkerless2022} provides the first LiDAR motion capture benchmark and baseline with a GCN-based inverse kinematic solver. However, its ideal collection environment limits its application scenarios. To tackle this, \cite{fanLiDARHMR3DHuman2025,anPretrainingDensityAwarePose2025,qu2024conditional} ultilize multi-level or generative models to deal with point clouds of different sparsity, \cite{zhangNeighborhoodEnhanced3DHuman2024} captures extra environmental clues from points in the background to enhance the pose learning. On the other hand, LiveHPS \cite{renLiveHPSLiDARbasedScenelevel2024} and LiveHPS++ \cite{renLiveHPSRobustCoherent2024} achieves robust motion tracking by exploiting temporal and spatial coherence priors of the point cloud frames. There are also some works that customize modeling for specific scenarios \cite{yanCIMI4DLargeMultimodal2023,daiHSC4DHumancentered4D2022,daiHiSC4DHumanCenteredInteraction2024}. 

Although these methods have achieved impressive performance, they primarily exploit action priors from specific point cloud patterns, which limits their ability to handle missing frames in the input. In this work, we address this issue by leveraging the intrinsic properties of motion itself. Prior studies \cite{maProgressivelyGeneratingBetter2022a,linProgressivePretextTask2024} suggest that progressively regressing motions can improve accuracy and smoothness, but such approaches are often slow due to their iterative inference \cite{qu2025end, qu2025robust}. To overcome this limitation, we design a single-stage Transformer architecture. By regulating information flow through causal mask and selectively processing informative point cloud features, the model is able to reconstruct multi-level human motions in a single forward pass.

\subsection{Motion Representation}

The representation of motion plays a critical role in determining feature quality and the final performance of pose estimation. A straightforward way avoids additional processing by directly inferring the pose and body shape for each frame, such as methods that directly regress the SMPL parameters of the human body~\cite{kanazawaEndtoendRecoveryHuman2018,zhangPyMAF3DHuman2021,liCLIFFCarryingLocation2022,sunMonocularOneStageRegression2021,dwivediPOCO3DPose2024,kocabasVIBEVideoInference2020,goelHumans4DReconstructing2023}. On the other hand, with the recent advances in prior distribution learning using VQ-VAE~\cite{oordNeuralDiscreteRepresentation2018}, some HPS methods attempt to obtain latent representations that are more amenable to Transformer learning, as in pose modeling approaches~\cite{ficheVQHPSHumanPose2025,dwivediTokenHMRAdvancingHuman2024,ficheMEGAMaskedGenerative2025,gengHumanPoseCompositional2023a}. However, most of these approaches encode only at the pose level and lack temporal modeling. Another line of work directly models motion by incorporating richer kinematic cues, such as velocity and ground contact~\cite{jangMOVINRealtimeMotion2023}, or by compressing frames into a single latent representation~\cite{guoTM2TStochasticTokenized2022,tevetHumanMotionDiffusion2022a}. Nevertheless, their frame-by-frame inference paradigm tends to accumulate errors over time.

We aim to address this issue through progressive deduction at the joint trajectory level. Recent studies have explored learning multi-stage residual latent representations~\cite{gongCARPVisuomotorPolicy2025,guoMoMaskGenerativeMasked2024}, while frequency-decomposition-based methods~\cite{zhongFreqPolicyFrequencyAutoregressive2025,sunHUMOFHumanMotion2025} further enhance interpretability and reduce dependence on VQ-VAE. However, since the additional signal components produced at each stage are orthogonal, errors in earlier signals are difficult to correct. Meanwhile, motion inbetweening tasks~\cite{studerFactorizedMotionDiffusion2024} highlight the quadratic curves are easy to adjust and correct in modeling human motion. Inspired by these observations, we propose a multi-level Bézier curve–based motion representation and degradation strategy, which not only generates hierarchical motion representations but also preserves a high degree of similarity with the original joint trajectories.
\section{Methodology}
\begin{figure*}[t]
    \centering
    \includegraphics[width=.95\textwidth]{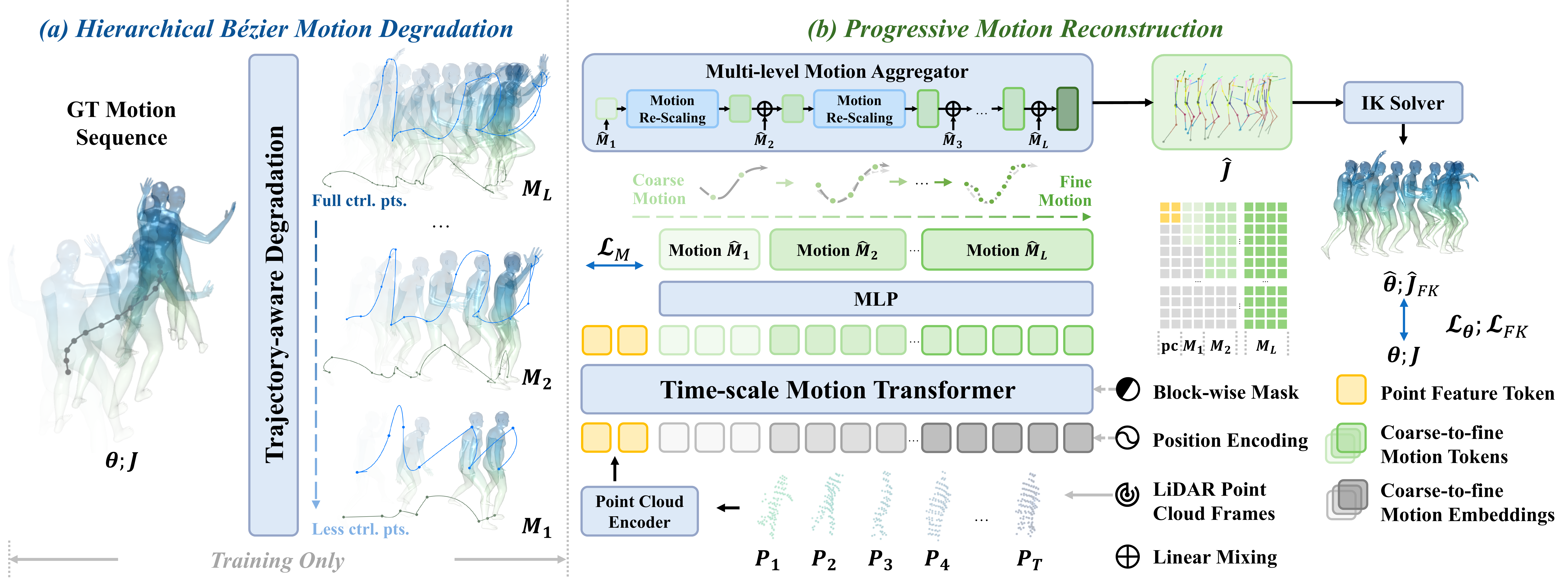}
    \begin{subfigure}[t]{0.4\textwidth}\phantomsubcaption\label{fig:arch_bmd}
    \end{subfigure}
    \begin{subfigure}[t]{0.4\textwidth}\phantomsubcaption\label{fig:arch_pmr}
    \end{subfigure}
    \vspace{-.5em}
    \caption{The pipeline of our proposed BMLiCap framework. During training, we first apply the Bézier motion degradation module to generate multi-level motion representations. Then, the progressive motion reconstruction module reconstructs the motion in a coarse-to-fine manner, where a Time-scale Motion Transformer (TMT) predicts motion curves at different temporal scales conditioned on LiDAR features, and a Multi-level Motion Aggregator (MMA) fuses these multi-scale cues to produce the final fine-grained motion.}
    \label{fig:arch}
\end{figure*}

In this section, we introduce \textbf{BMLiCap}, a novel framework for LiDAR-based 3D human motion capture. Our key idea is to leverage a Bézier multi-level motion representation to progressively reconstruct human motion in a coarse-to-fine manner. To achieve this, as illustrated in Fig.~\ref{fig:arch}, we design two major components:  
\textbf{(a)} A trajectory-aware Bézier motion degradation module that generates learning-friendly multi-level motion representations during training.  
\textbf{(b)} A progressive motion reconstruction module that employs a Time-scale Motion Transformer to jointly construct motion trajectories across multiple temporal scales in a single forward pass, while a Multi-level Motion Aggregator integrates these multi-scale cues to produce the final reconstructed motion.

\noindent\textbf{Problem Definition.} Given a sequence of LiDAR point clouds $\mathcal{P}=\{\mathbf{P}_t\in \mathbb{R}^{N\times3}\}_{t=0}^{T-1}$ captured over $T$ frames, where $N$ is the number of points, our goal is to estimate the corresponding 3D human motion $\mathcal{M}=\{\boldsymbol{\theta}_t\in\mathbb{R}^{K\times3},\mathbf{J}_t\in\mathbb{R}^{K\times 3}\}_{t=0}^{T-1}$, where $K$ is the number of body joints, $\mathbf{J}_t$ is the joint locations and $\boldsymbol{\theta}_t$ the pose parameters of a standard SMPL model.

\subsection{Hierarchical Bézier Motion Degradation}
\label{sec:bezier_degradation}
\noindent\textbf{Initial Bézier Fitting.}
To obtain the finest-grained motion representation and prepare for subsequent stages, we first fit the original joint trajectories with Bézier curves. To ensure the smoothness of the fitted curves, we enforce $C^1$ continuity at each control point. Specifically, given the trajectory of joint $k$ across $T$ frames' time, denoted as $\mathbf{J}^{(k)}\in \mathbb{R}^{T\times 3}$, we treat each $\mathbf{J}^{(k)}_t$ as an anchor point and construct $T-1$ cubic Bézier curves:
\begin{equation}
    \begin{aligned}
        \mathcal{B}^{(k)}_t(u) &= (1-u)^3\mathbf{J}^{(k)}_t + 3(1-u)^2u\mathbf{C}^{(k)}_{t,2} \\
         &+ 3(1-u)u^2\mathbf{C}^{(k)}_{t+1,1} + u^3\mathbf{J}^{(k)}_{t+1}, \\
         \overrightharpoon{\mathbf{C}^{(k)}_{t,1}\mathbf{J}^{(k)}_{t}}&= \overrightharpoon{\mathbf{J}^{(k)}_{t}\mathbf{C}^{(k)}_{t,2}}, \quad t=0,\dots,T-2,
    \end{aligned}
\end{equation}
where $\mathbf{C}^{(k)}_{t,1},\mathbf{C}^{(k)}_{t,2}\in \mathbb{R}^{3}$ are the backward and forward control points, $u\in[0,1]$ is the curve parameter. By setting initial acceleration $\ddot{\mathcal{B}}^{(k)}_0(0)=0$, we can solve all control points $\mathbf{C}^{(k)}$ by Thomas algrithms. Then, we have the finest cubic Bézier chain $\{{\mathbf{J}}^{(k)}_i,{\mathbf{C}}^{(k)}_{i,1},{\mathbf{C}}^{(k)}_{i,2}\}_{i=0}^{T-1}$ representing the original joint trajectory. 

\noindent\textbf{Trajectory-Aware Degradation (TAD). }
To obtain coarse-to-fine motion representations that are both easy for network learning and suitable for loss design, we aim to progressively reduce the temporal resolution of motions while preserving the overall dynamic trends. To this end, we propose a hierarchical motion degradation strategy. Specifically, we perform trajectory downsampling by selecting anchors and adjusting the control points of the initial Bézier curves. 

\begin{figure}[t]
    \centering
    \includegraphics[width=0.8\linewidth]{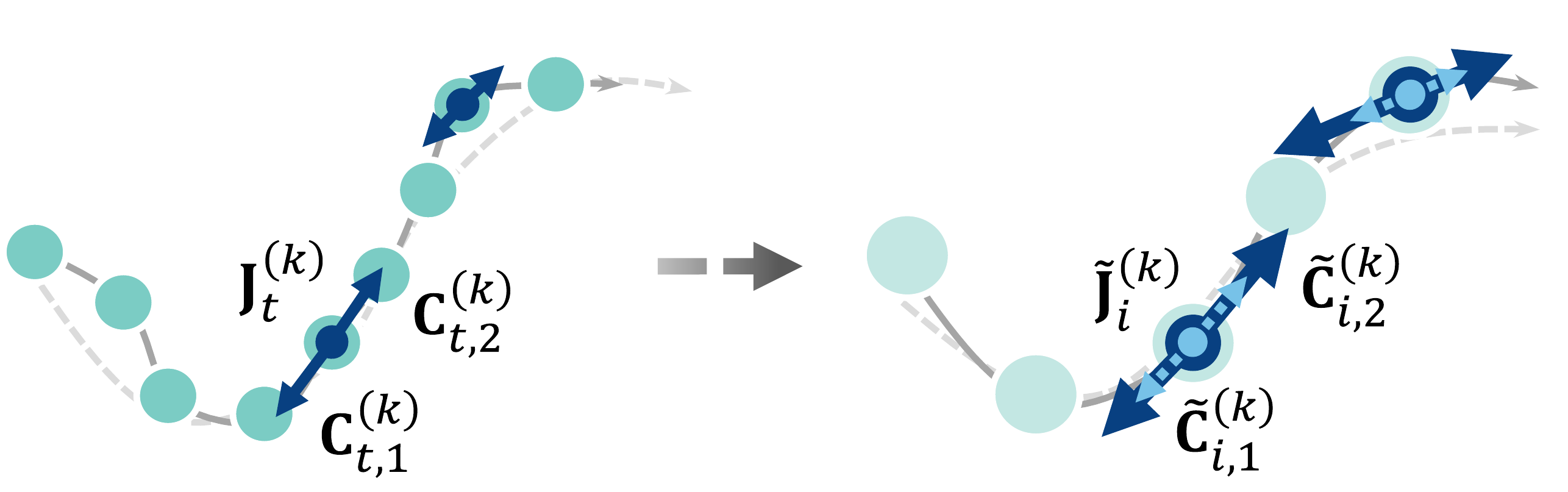}
    \caption{A demonstration of trajectory-aware Bézier degradation, we not only resample the control points but also adjust their lengths to better fit the finest curve.}
    \label{fig:tad}
\end{figure}

As shown in Fig.\ref{fig:tad}, given a downsampling step size $s$, the length of the new trajectory is degraded to $M_s=\lceil{T}/{s}\rceil$. We first uniformly sample $M_s$ time indices $\mathcal{T}_s=\{t_0=0, t_1=s, \dots,t_{M_s-1}=T-1\}$ from the original time steps. Then, we extract the corresponding joint positions $\widetilde{\mathbf{J}}^{(k)}_i = \mathbf{J}^{(k)}_{t_i}$ as the new anchor points. To better preserve the motion dynamics, we first extract the unit tangent vectors $\widehat{\mathbf{d}}^{(k)}_i$ at each $\widetilde{\mathbf{J}}^{(k)}_i$ from the finest curve as:
\begin{equation}
    \widehat{\mathbf{d}}^{(k)}_i = \widetilde{\mathbf{J}}^{(k)}_{i}\mathbf{C}^{(k)}_{t_i,1}/\|\widetilde{\mathbf{J}}^{(k)}_{i}\mathbf{C}^{(k)}_{t_i,1}\|_2.
\end{equation}
Then, the new control points are defined as:
\begin{equation}
    \widetilde{\mathbf{C}}^{(k)}_{i,1} = \widetilde{\mathbf{J}}^{(k)}_i - \ell_{i,1}\,\widehat{\mathbf{d}}^{(k)}_i, \quad
    \widetilde{\mathbf{C}}^{(k)}_{i,2} = \widetilde{\mathbf{J}}^{(k)}_{i} + \ell_{i,2}\,\widehat{\mathbf{d}}^{(k)}_{i}.
\end{equation} 
This form the new Bézier chain $\{\widetilde{\mathbf{J}}^{(k)}_i,\widetilde{\mathbf{C}}^{(k)}_{i,1},\widetilde{\mathbf{C}}^{(k)}_{i,2}\}_{i=0}^{M_s-1}$. To better approximate the original motion dynamics, we solve the optimal lengths $\ell_i$ by:
\begin{equation}
    \min _{\left\{\ell_{i,2},\ell_{i+1,1}\right\}} \sum_m\left\|\tilde{\mathcal{B}}_i^{(k)}\left(u_{i, m}\right)-\mathbf{Y}_{i, m}^{(k)}\right\|_2^2, i=0,\dots,M_s-1
\end{equation}
where $\tilde{\mathcal{B}}_i^{(k)}$ is the degraded Bézier curve between $\widetilde{\mathbf{J}}^{(k)}_i$ and $\widetilde{\mathbf{J}}^{(k)}_{i+1}$, $\mathbf{Y}_{i, m}^{(k)}$ are the original joint positions sampled from the finest curve within the time segment $[t_i,t_{i+1}]$, $u_{i,m}$ are the corresponding curve parameters. This least squares problem has a closed-form solution. The detailed procedure is described in the appendix.

By using different step sizes $\mathcal{S}=\{s_1,s_2,\dots,s_L\}$, we can obtain a series of degraded Bézier chains at multiple temporal scales. These chains are then packed to get the multi-level motion representation $\{\mathbf{M}_l\in\mathbb{R}^{M_{s_l}\times K\times 9}\}_{l=1}^L$. Generally, we set $s_l > s_{l+1}$ and  $s_L=1$ to form a coarse-to-fine hierarchy while retaining the finest motion representation.

During training, we supervise the network to predict these multi-level motion representations, enabling the model with coarse-to-fine motion reconstruction capability.

\subsection{Progressive Motion Reconstruction}
\noindent\textbf{Overall Architecture.} To effectively leverage the multi-level motion representations generated by the Bézier degradation module, inspired by the previous works \cite{gongCARPVisuomotorPolicy2025,tianVisualAutoregressiveModeling2024a}, we design a progressive motion reconstruction network, which consists of a Time-scale Motion Transformer (TMT) and a Multi-level Motion Aggregator (MMA). The TMT is responsible for reconstructing motion trajectories at different temporal scales conditioned on LiDAR features, while the MMA integrates these multi-scale cues to produce the final fine-grained motion sequence.

\noindent\textbf{LiDAR Feature Extraction.} Given the input point cloud sequence $\mathcal{P}$, the per-frame point features are extracted using a PointNet++ \cite{qiPointNetDeepHierarchical2017} and map to the dimension through an MLP. Then, we have the observation condition $\mathbf{F}_\mathcal{P}\in\mathbb{R}^{T\times D}$ for subsequent reasoning.

\noindent\textbf{Time-scale Motion Transformer (TMT).} To model the information flow between LiDAR features and multi-level motion representations, we design a Time-scale Motion Transformer (TMT) using an encoder-only architecture. As shown in Fig.\ref{fig:arch_pmr}, the TMT treats each level of motion representation as a separate token sequence. Formally, given the initial multi-level motion embeddings $\{\mathbf{E}_l\in\mathbb{R}^{M_{s_l}\times D}\}_{l=1}^{L}$ and LiDAR features $\mathbf{F}_\mathcal{P}$, the TMT jointly models their interactions and outputs the reconstructed motion curves at each temporal scale:
\begin{equation}
    \{\widehat{\mathbf{M}}_l\}_{l=1}^{L} = \operatorname{MLP}\left(\operatorname{TMT}(\mathbf{F}_\mathcal{P}, \{\mathbf{E}_l\}_{l=1}^{L})\right).
\end{equation}

To regularize the information flow between different stages, we impose a block-wise causal mask \cite{gongCARPVisuomotorPolicy2025,tianVisualAutoregressiveModeling2024a} on the self-attention layers, allowing each motion token to only attend to all tokens from coarser levels and all point feature tokens. This design ensures that coarse-level motion trends can effectively guide the refinement of finer motions, while also leveraging the available visual cues from the LiDAR features. 

\noindent\textbf{Multi-level Motion Aggregator (MMA).} To effectively fuse the multi-scale motion representations predicted by the TMT, we introduce a Multi-level Motion Aggregator (MMA). Specifically, the MMA employs a reduction mechanism to integrate the motion representations from different temporal scales progressively: 
\begin{equation}
    \widehat{\mathbf{M}}_{l+1}' =
    \begin{cases} 
        \operatorname{MLP}(\operatorname{Resample}(\widehat{\mathbf{M}}_{l}'), \widehat{\mathbf{M}}_{l+1}), & l=2,\dots,L-1 \\
        \widehat{\mathbf{M}}_{l}, & l=1
    \end{cases}   
\end{equation}
where $\operatorname{Resample}(\cdot)$ upsamples the coarser motion representation to match the length of the finer one using the predicted Bézier curve parameters, and $\operatorname{MLP}(\cdot)$ fuses the two representations. Finally, we take the location parts of the finest fused motion representation $\widehat{\mathbf{M}}_{L}'$ as the final joint location predictions $\{\widehat{\mathbf{J}}_t\}_{t=0}^{T-1}$.

\noindent\textbf{Inverse Kinematic Solver.} Since our model predicts joint locations, following \cite{liLiDARCapLongrangeMarkerless2022,renLiveHPSRobustCoherent2024}, we adopt an STGCN-based \cite{hanSTGCNSpatialTemporalAware2020} inverse kinematic (IK) solver to convert the estimated joint positions $\{\widehat{\mathbf{J}}_t\}_{t=0}^{T-1}$ into SMPL pose parameters $\{\widehat{\boldsymbol{\theta}}_t\}_{t=0}^{T-1}$. Then, we calculate the SMPL forward kinematics using the predicted pose parameters:
\begin{equation}
    \widehat{\mathbf{J}}_{t;\text{FK}} = \operatorname{SMPL}(\widehat{\boldsymbol{\theta}}_t,\beta).
\end{equation}

\noindent\textbf{Loss Functions.} To supervise the motion reconstruction process, we apply multi-level supervision on the predicted Bézier motion representations:
\begin{equation}
    \mathcal{L}_{M} = \sum_{l=1}^{L} \frac{1}{M_{s_l}}\left\|\widehat{\mathbf{M}}_l - \mathbf{M}_l\right\|_F^2.
\end{equation}
Besides, we apply the parameter loss and forward kinematic loss to supervise the learning of the IK solver:
\begin{equation}
    \begin{aligned}
        \mathcal{L}_{\theta} &= \frac{1}{KT}\sum_{t=0}^{T-1} \left\|\boldsymbol{\theta}_t - \widehat{\boldsymbol{\theta}}_t\right\|_F^2,\\
        \mathcal{L}_\text{FK} &= \frac{1}{KT}\sum_{t=0}^{T-1} \left\|\mathbf{J}_t - \widehat{\mathbf{J}}_{t;\text{FK}}\right\|_F^2.
    \end{aligned}
\end{equation}
Altogether, the overall training loss is defined as:
\begin{equation}
    \mathcal{L} = \lambda_M \mathcal{L}_M + \lambda_{\theta} \mathcal{L}_{\theta} + \lambda_\text{FK} \mathcal{L}_\text{FK},
\end{equation}
where $\lambda_M$, $\lambda_{\theta}$ and $\lambda_\text{FK}$ are the weights for each loss term.
\section{Experiments}
In this section, we organize comprehensive experiments to evaluate the effectiveness of our proposed BMLiCap framework. We compare our method with state-of-the-art approaches quantitatively and qualitatively on four LiDAR-based motion capture benchmarks, including LiDARHuman26M \cite{liLiDARCapLongrangeMarkerless2022}, FreeMotion \cite{renLiveHPSLiDARbasedScenelevel2024}, NoiseMotion \cite{renLiveHPSRobustCoherent2024}, and SLOPER4D \cite{daiSLOPER4DSceneAwareDataset2023}. These datasets cover diverse scenarios, from controlled indoor environments to complex outdoor scenes with varying levels of occlusion and noise. We also conduct ablation studies on LiDARHuman26M to analyze the contributions of different components in our framework.

\begin{table*}[ht]
\centering
\caption{Comparison with state-of-the-art methods on four mainstream benchmarks. ``M." and ``S." denote methods that utilize multiple frames or a single frame as input, respectively. ``†" indicates an 32-frame variant of our method. Best results are marked by \colorbox{gray}{\textbf{bold blue}}, second best by \colorbox{lightgray}{light blue}. AE is not available for single-frame methods.}
\label{tab:performance_comparison}
\resizebox{\linewidth}{!}{
\begin{tabular}{p{3.5cm}|c|ccc|ccc|ccc|ccc}
\toprule
\multicolumn{1}{c|}{\multirow{2}{*}{\textbf{Method}}} & \multirow{2}{*}{\textbf{Type}} & \multicolumn{3}{c|}{\textbf{LiDARHuman26M}} & \multicolumn{3}{c|}{\textbf{FreeMotion}} & \multicolumn{3}{c|}{\textbf{NoiseMotion}} & \multicolumn{3}{c}{\textbf{SLOPER4D}} \\ 
\multicolumn{1}{c|}{} &  & JPE↓ & VPE↓ & AE↓ & JPE↓ & VPE↓ & AE↓ & JPE↓ & VPE↓ & AE↓ & JPE↓ & VPE↓ & AE↓ \\ \midrule
MOVIN~\cite{jangMOVINRealtimeMotion2023} & M. & - & - & - & - & - & - & - & - & - & 123.8 & 146.3 & 106.3 \\
LiDAR-HMR~\cite{fanLiDARHMR3DHuman2025} & S. & 76.2 & 102.5 & - & 106.7 & 130.7 & - & 50.4 & 65.2 & - & 47.7 & 49.7 & - \\
LiDARCap~\cite{liLiDARCapLongrangeMarkerless2022} & M. & 79.3 & 101.6 & 45.2 & 86.3 & 104.2 & 62.8 & 52.6 & 64.7 & 42.5 & 71.6 & 84.2 & 40.1 \\
LIP$^*$~\cite{renLiDARaidInertialPoser2023} & M. & 75.7 & 96.6 & 32.8 & 85.5 & 90.8 & 61.6 & 62.4 & 78.0 & 25.3 & 60.1 & 74.9 & 61.6 \\
NE-LiDARCap$^*$~\cite{zhangNeighborhoodEnhanced3DHuman2024} & M. & 76.8 & 97.1 & 31.6 & 62.5 & 75.8 & 29.8 & 48.8 & 60.4 & 27.9 & 96.8 & 113.9 & 38.6 \\
LiveHPS~\cite{renLiveHPSLiDARbasedScenelevel2024} & M. & 71.9 & 92.1 & 34.1 & 69.4 & 83.9 & 69.9 & 48.4 & 60.4 & 57.8 & 53.4 & 63.2 & 58.8 \\
LiveHPS++~\cite{renLiveHPSRobustCoherent2024} & M. & - & - & - & 61.9 & 75.3 & 54.2 & \cellcolor{gray}\textbf{34.0} & \cellcolor{gray}\textbf{42.8} & 34.8 & 42.7 & 50.6 & 43.4 \\
\cellcolor{lightgray}\textbf{BMLiCap (Ours)} & \cellcolor{lightgray}M. & \cellcolor{lightgray}70.1 & \cellcolor{lightgray}89.5 & \cellcolor{lightgray}31.2 & \cellcolor{lightgray}49.6 & \cellcolor{lightgray}60.3 & \cellcolor{lightgray}27.1 & \cellcolor{gray}\textbf{34.0} & \cellcolor{gray}\textbf{42.8} & \cellcolor{lightgray}24.1 & \cellcolor{lightgray}39.7 & \cellcolor{lightgray}47.8 & \cellcolor{lightgray}22.3 \\
\cellcolor{gray}\textbf{BMLiCap (Ours)} † & \cellcolor{gray}M. & \cellcolor{gray}\textbf{66.8} & \cellcolor{gray}\textbf{85.4} & \cellcolor{gray}\textbf{28.8} & \cellcolor{gray}\textbf{47.2} & \cellcolor{gray}\textbf{59.0} & \cellcolor{gray}\textbf{22.5} & \cellcolor{lightgray}36.9 & \cellcolor{lightgray}47.0 & \cellcolor{gray}\textbf{23.8} & \cellcolor{gray}\textbf{36.5} & \cellcolor{gray}\textbf{44.2} & \cellcolor{gray}\textbf{13.6} \\ \bottomrule
\end{tabular}}
\vspace{-1em}
\end{table*}

\subsection{Implementation Details}
We build our proposed method base on PyTorch 2.3.1 with CUDA 11.8. We follow the baseline method LiDARCap \cite{liLiDARCapLongrangeMarkerless2022} on most modules and settings. The point cloud encoder is a PointNet++ \cite{qiPointNetDeepHierarchical2017}, pretrained on synthesized human instances \cite{anPretrainingDensityAwarePose2025}. The Time-scale Motion Transformer is a standard Transformer encoder with $n_\text{layer}=12, n_\text{dim}=512, n_\text{head}=16$. We test different $\mathcal{S}$ settings, and the results are shown in the ablation study. The weights for loss terms are set as $\lambda_M=0.5$, $\lambda_{\theta}=\lambda_\text{FK}=1.0$. We train our model using the AdamW \cite{loshchilovDecoupledWeightDecay2019} optimizer with a learning rate of $2.5\times10^{-4}$ for 50 epochs on $4\times$ NVIDIA RTX 4090 GPUs.

\subsection{Comparison Methods and Metrics}

We compare our BMLiCap with state-of-the-art (SOTA) LiDAR-based human motion capture approaches, including LiDAR-HMR \cite{liLiDARCapLongrangeMarkerless2022}, MOVIN \cite{jangMOVINRealtimeMotion2023}, LiDARCap \cite{liLiDARCapLongrangeMarkerless2022}, LIP \cite{renLiDARaidInertialPoser2023}, NE-LiDARCap \cite{zhangNeighborhoodEnhanced3DHuman2024}, LiveHPS \cite{renLiveHPSLiDARbasedScenelevel2024}, and LiveHPS++ \cite{renLiveHPSRobustCoherent2024}. For a fair comparison, we remove the auxiliary inputs for LIP (inertial) and NE-LiDARCap (3SN, 3BN). Following \cite{renLiveHPSLiDARbasedScenelevel2024}, we include three widely used metrics for evaluation: i) Mean Per Joint Position Error (MPJPE/JPE)$\downarrow$ in $mm$ , ii) Mean Per Vertex Position Error (MPVPE/PVE)$\downarrow$ in $mm$, and iii) Acceleration Error (Accel Err/AE)$\downarrow$ in $cm/s^2$. The MPJPE/MPVPE measure the relative position bias of the predicted joints, and the Accel Err quantifies the coherence and smoothness of the predicted motion.

\subsection{Quantitative Analysis}
\noindent\textbf{Overall Benchmark.}
As shown in Tab.~\ref{tab:performance_comparison}, across four mainstream benchmarks and three standard metrics, our method establishes a new state of the art. 
Using the default temporal window as in other methods, BMLiCap already achieves the best performance. 
Further increasing the window to 32 frames (BMLiCap$\dagger$) yields higher accuracy and better motion coherence. 
Specifically, on the challenging FreeMotion dataset, BMLiCap$\dagger$ achieves improvements of 14.7 MPJPE, 16.3 MPVPE, and 31.7 Accel Err over the previous best method LiveHPS++~\cite{renLiveHPSRobustCoherent2024}. 
On the other three datasets, our method consistently outperforms existing approaches by notable margins across all metrics. 
These results demonstrate the effectiveness of the proposed BMLiCap framework in capturing accurate and temporally coherent human motion from LiDAR data. 
In particular, we observe that on NoiseMotion, a shorter window performs better. 
We attribute this to the fact that the dataset exhibits more viewport jumps, causing longer temporal windows to aggregate more corrupted or slightly misaligned location annotations, which degrades JPE/ VPE, as these metrics are sensitive to alignment.

\begin{figure}[t]
\centering
\includegraphics[width=\linewidth]{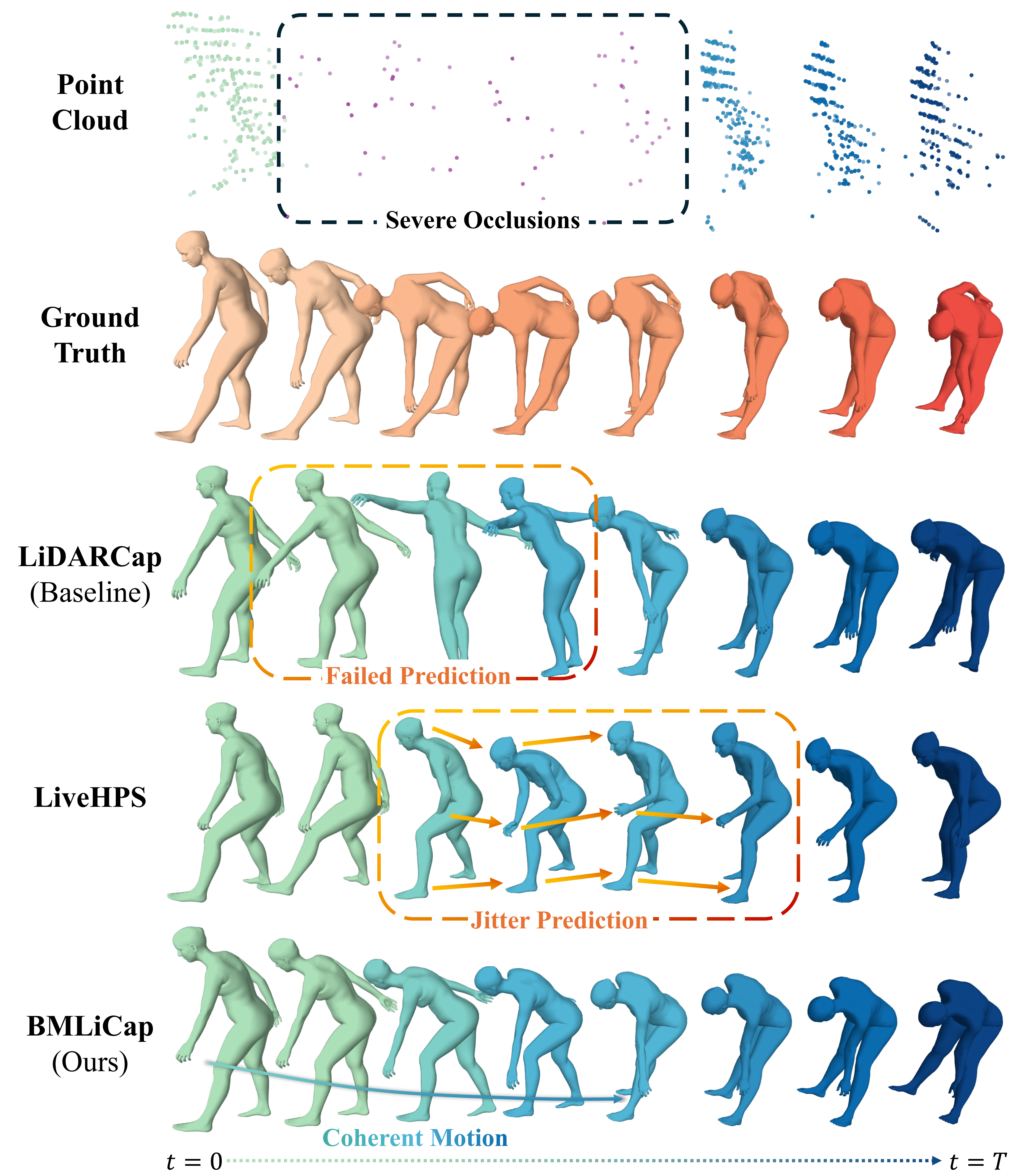}
\caption{Sequential visualization. Even under severe occlusion, BMLiCap provides coherent, accurate estimations, while other methods produce jittery/failed results.}
\label{fig:seq_vis}
\vspace{-1em}
\end{figure}

\begin{figure*}[t]
\centering
\makebox[\textwidth][c]{
\begin{minipage}[t]{.64\textwidth}%
  \vspace{0pt}
  \begingroup
  \newlength{\capwlocal}
  \setlength{\capwlocal}{0.015\linewidth}
  \setlength{\tabcolsep}{6pt}
  \renewcommand{\arraystretch}{1.1}
  \noindent\hspace*{4\capwlocal}%
  \footnotesize
  \begin{tabularx}{.92\linewidth}{*{7}{>{\centering\arraybackslash}X}}
    PC & GT & Ours & NELiCap & LiCap & LiveHPS & LIP \\
  \end{tabularx}

  \begin{tabular}{@{}m{2.2\capwlocal}m{\dimexpr\linewidth-\capwlocal\relax}@{}}
    \centering\raisebox{2.0em}{\rotatebox{90}{\footnotesize LiDARH26M~\cite{liLiDARCapLongrangeMarkerless2022}}} &
    \includegraphics[width=.92\linewidth]{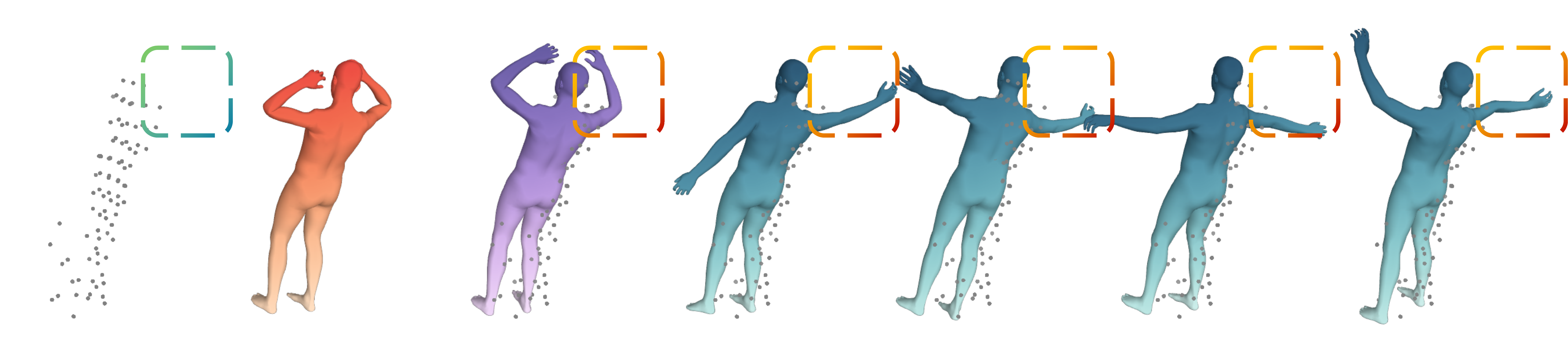} \\[-2.3em]
    \centering\raisebox{-1em}{\rotatebox{90}{\footnotesize \quad FreeMotion~\cite{renLiveHPSLiDARbasedScenelevel2024}}} &
    \includegraphics[width=.92\linewidth]{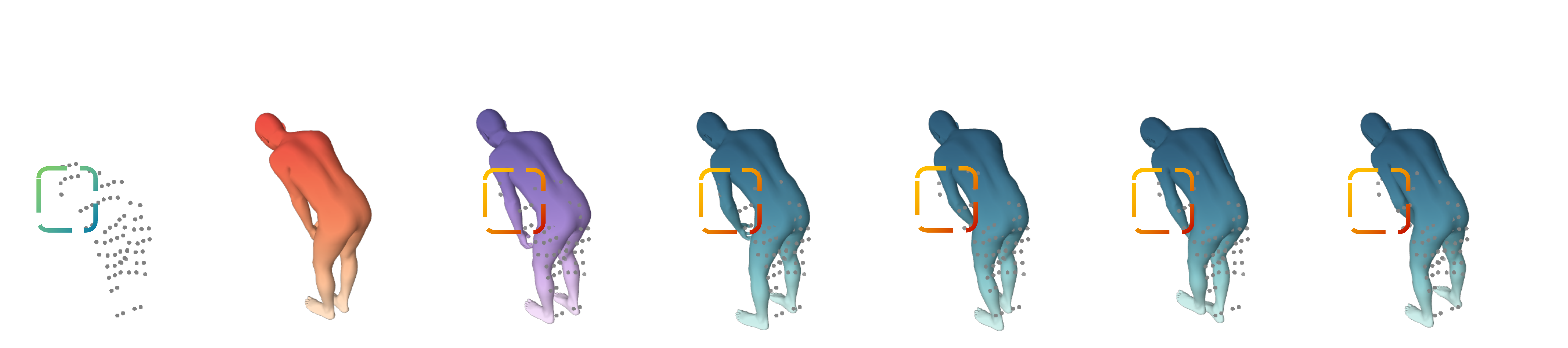} \\[-1em]
    \centering\raisebox{-.5em}{\rotatebox{90}{\footnotesize NoiseMotion~\cite{renLiveHPSRobustCoherent2024}}} &
    \includegraphics[width=.92\linewidth]{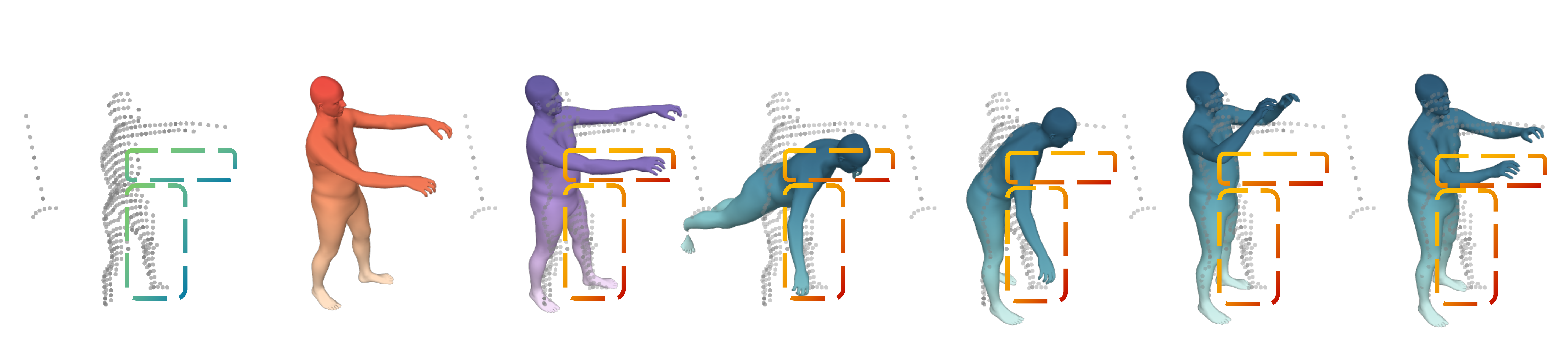} \\[-1.2em]
    \centering\raisebox{-.8em}{\rotatebox{90}{\footnotesize SLOPER4D~\cite{daiSLOPER4DSceneAwareDataset2023}}} &
    \includegraphics[width=.92\linewidth]{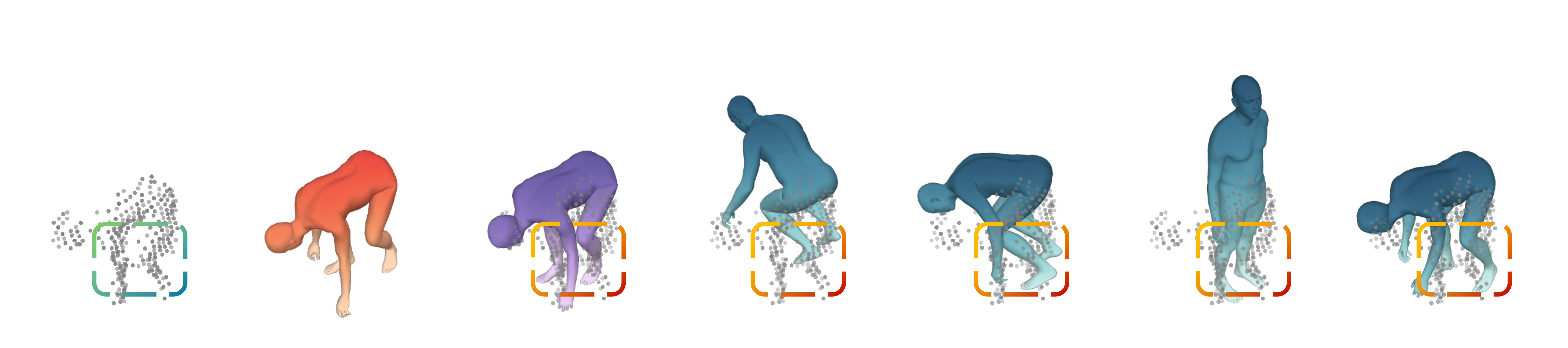} \\
  \end{tabular}

  \caption{Single frame visual comparisons. On samples with special motion or severe occlusion, our method can compensate for defects, producing stable and coherent results.}
  \label{fig:example_vis_all}
  \endgroup
\end{minipage}%
\hspace{.02\linewidth}
\nextfloat
\begin{minipage}[t]{.34\textwidth}%
  \vspace{0pt}
  \centering
  \captionsetup{font=small}
  \captionsetup[sub]{aboveskip=2pt, belowskip=2pt}

  \begin{subfigure}[t]{\linewidth}
    \centering
    \includegraphics[width=.95\linewidth]{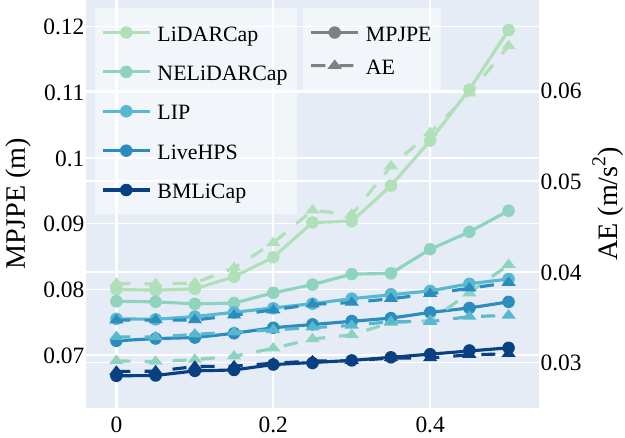}
    \caption{continuous frame loss.}
    \label{fig:mask_continuous}
  \end{subfigure}

  \vspace{.7em}


  \begin{subfigure}[t]{\linewidth}
    \centering
    \includegraphics[width=.95\linewidth]{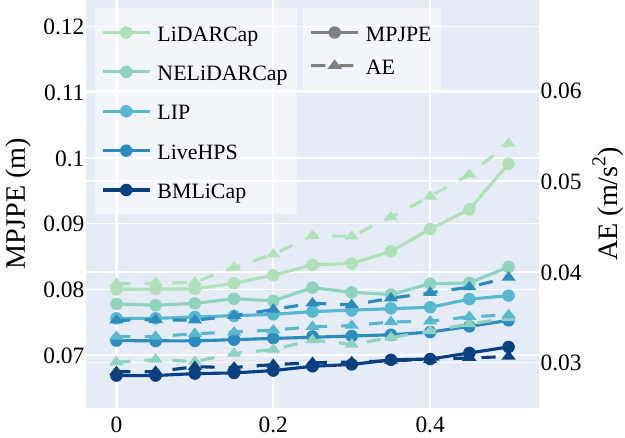}
    \caption{random frame loss.}
    \label{fig:mask_random}
  \end{subfigure}

  \vspace{-.4em}
  \caption{Stability test under different point cloud frame masking policy and ratio.}
  \label{fig:mask_tests}
\end{minipage}%
}
\vspace{-1.0em}
\end{figure*}

\begin{figure*}[t]
  \centering
  \vspace{-3.5em}
  \begin{subfigure}[t]{.31\linewidth}
    \includegraphics[width=\textwidth]{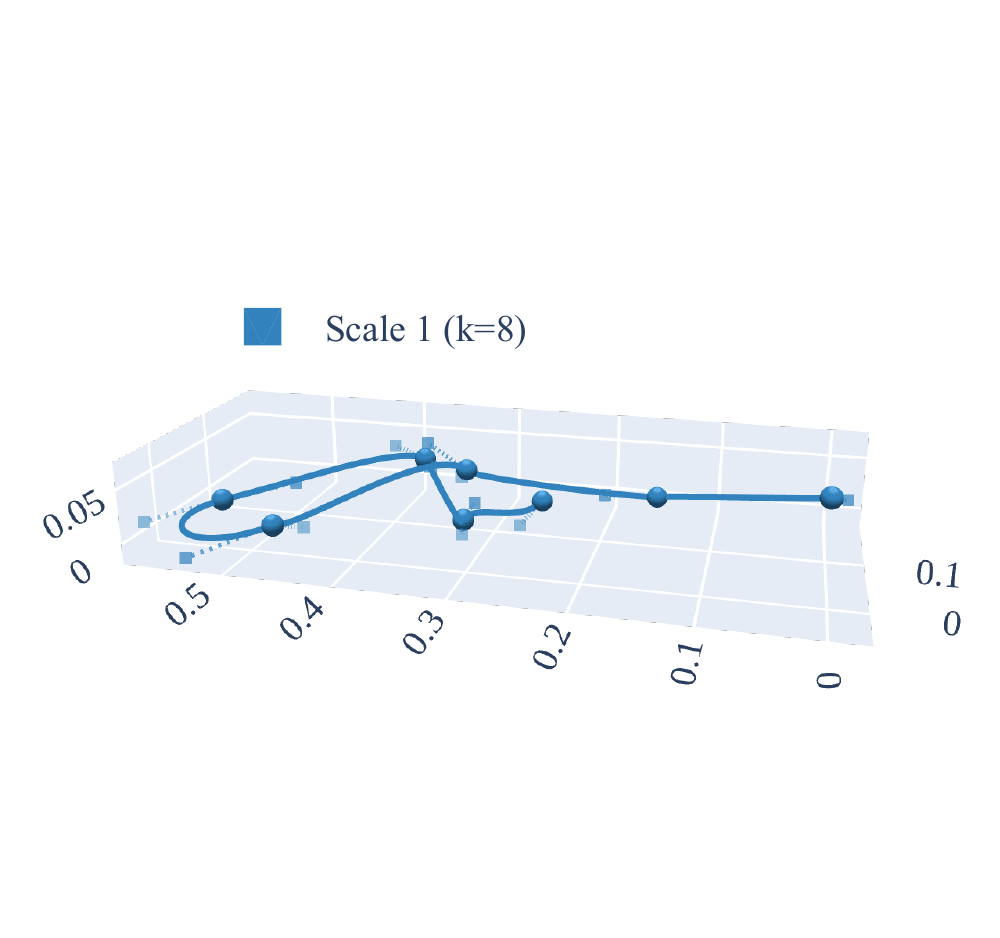}
  \end{subfigure}
  \hfill
  \begin{subfigure}[t]{.31\linewidth}
    \includegraphics[width=\textwidth]{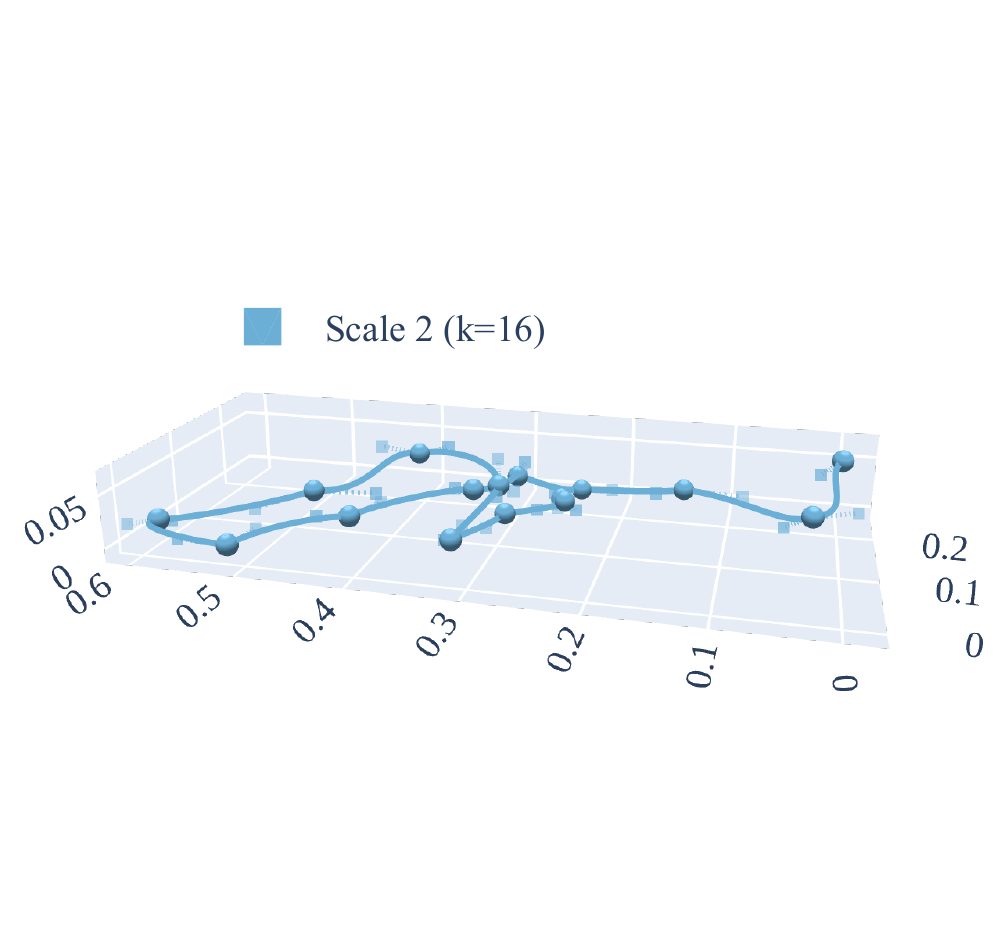}
  \end{subfigure}
  \hfill
  \begin{subfigure}[t]{.31\linewidth}
    \includegraphics[width=\textwidth]{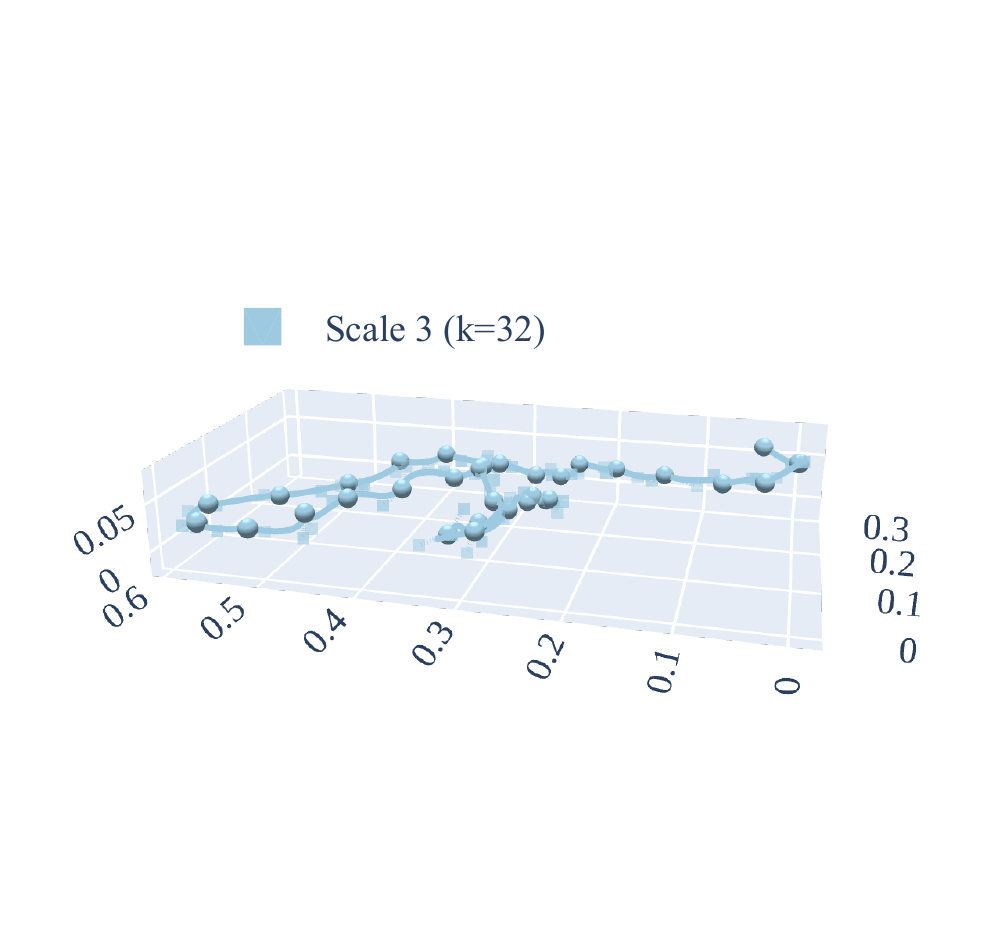}
  \end{subfigure}
  \vspace{-4em}
  \caption{Visualization of the predicted intermediate Bézier curves from each level of BMLiCap ($\widehat{\mathbf{M}}_1$, $\widehat{\mathbf{M}}_2$, $\widehat{\mathbf{M}}_3$). The curves evolve from an initial trend to richer details, matching the desired target representation.}
  \label{fig:seq_bezire}
  \vspace{-1em}
\end{figure*}

\begin{table}[t]
\caption{Comparison with other motion representations on the LiDARHuman26M benchmark.}
\label{fig:performance_rep}
\centering
\resizebox{\linewidth}{!}{
\begin{tabular}{L{3.1cm}|M{1.45cm}M{1.45cm}M{1.45cm}}
\toprule
\textbf{Method} & \textbf{MPJPE} & \textbf{MPVPE} & \textbf{Accel Err} \\ \midrule
Frequency-DCT \cite{zhongFreqPolicyFrequencyAutoregressive2025}  & 76.4 & 97.8 & 35.4 \\
VAE-smooth \cite{zhangLearningMotionPriors2021}   & 78.2 & 100.1 & 36.8 \\
Linear  & 75.7 & 96.3 & 35.5 \\
B-Spline   & 70.5 & 90.4 & 30.0 \\
\cellcolor{gray}\textbf{Bézier+TAD (Ours)}  & \cellcolor{gray}\textbf{66.8} & \cellcolor{gray}\textbf{85.4} & \cellcolor{gray}\textbf{28.8} \\ \bottomrule
\end{tabular}
}
\vspace{-1em}
\end{table}

\noindent\textbf{Motion Representation.}
To verify the effectiveness of the proposed motion representation, we replace the representation associated with $\mathcal{L}_M$ with alternative forms. As shown in Tab.~\ref{fig:performance_rep}, our method achieves better performance than simply using a linear or frequency-based degradation~\cite{zhongFreqPolicyFrequencyAutoregressive2025}. We attribute these gains to the $C^1$-continuous Bézier parameterization, which reduces velocity kinks and acceleration spikes. In contrast, linear interpolation only guarantees $G^0$ continuity (leading to speed discontinuities), B-Spline and VAE representation oversmooth the motion, and frequency decomposition using the discrete cosine transform (DCT) tends to introduce phase lag and ringing under aggressive degradation. The proposed TAD policy further adjusts the parameters of curve segments, balancing denoising and fidelity. This explains the notable reduction in Accel Err and the consistent improvements in MPJPE/MPVPE.

\noindent\textbf{Frame-Drop Robustness.}
To evaluate the robustness of our method under severe occlusions, we conduct stability tests by randomly removing a certain fraction of LiDAR point-cloud frames during inference. The points in the removed frames are replaced with 90\% meaningless placeholders for padding. As shown in Fig.~\ref{fig:mask_continuous}, our method maintains stable performance even when up to 50\% of the input frames are missing, demonstrating a strong ability to leverage coarse-level motion representations to compensate for observation gaps. This robustness is crucial for real-world applications where occlusions and unstable human tracking are common.

\subsection{Qualitative Analysis}
\noindent\textbf{Static Visual Comparisons.}
We present qualitative comparisons with state-of-the-art methods on various benchmarks in Fig.~\ref{fig:example_vis_all}. 
The selected samples all exhibit severe occlusions of body parts or unusual movements, particularly of the upper limbs. 
Specifically, on LiDARHuman26M, we showcase a sample in which the arm intermittently appears and disappears due to self-occlusion caused by body motion, where all other methods produce noticeably jittery predictions. 
We also present samples that involve complex pose changes, where neither LiDARCap nor LiveHPS can reliably capture the underlying motion. 
In contrast, BMLiCap consistently produces more accurate and temporally coherent motion reconstructions. 
The coarse-level motion representations enable our model to infer plausible human poses according to the learned kinematic priors, resulting in visually appealing and realistic motion sequences.

\noindent\textbf{Sequential Visual Comparisons.}
To verify that the model's coarse trajectories match the training targets, we visualize the multi-level trajectories in Fig.~\ref{fig:seq_bezire}, illustrating the progressive refinement process. 
We observe that the curves at the coarse stage capture the overall motion trend, while the curves at the fine stage add more local detail. 
Further, we also present an example with severe input occlusions (50\% missing) in Fig.~\ref{fig:seq_vis}. 
BMLiCap provides coherent and accurate estimates, whereas other methods suffer from noticeable jitter or even complete failure.

\begin{figure}
  \centering
  \begin{subfigure}[t]{\linewidth}
    \centering
    \includegraphics[width=\linewidth]{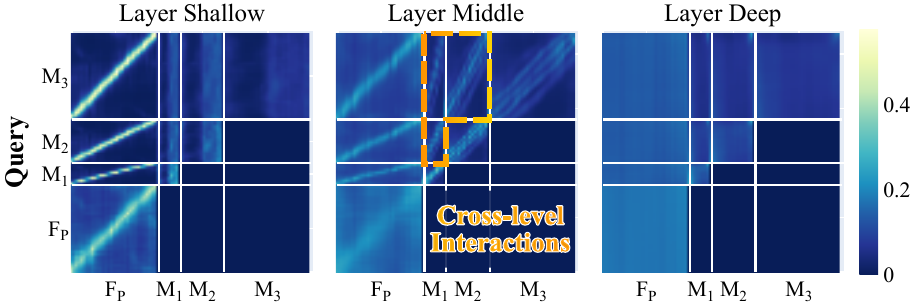}
    \caption{Normal sequence.}
    \label{fig:attn_map_ez}
  \end{subfigure}
  
  \vspace{.5em}
  
  \begin{subfigure}[t]{\linewidth}
    \centering
    \includegraphics[width=\linewidth]{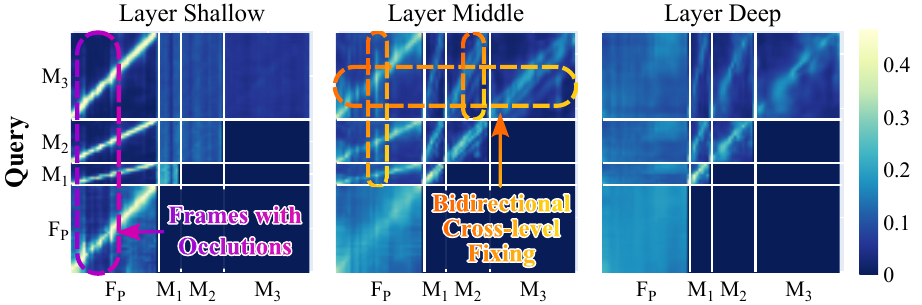}
    \caption{Heavy occluded sequence.}
    \label{fig:attn_map_hd}
  \end{subfigure}
  \vspace{-.6em}
  \caption{Attention map of different occlusion levels. The attention masking triggers cross-level interaction to fix missing frames.}
  \label{fig:attn_map}
\end{figure}

\noindent\textbf{Attention Maps in TMT.} To further understand how our model leverages multi-stage information, we visualize the attention maps from the Time-scale Motion Transformer for sequences with different occlusion levels in Fig.\ref{fig:attn_map}. In the normal sequence (Fig.\ref{fig:attn_map_ez}), attention between different stages tends to be diagonalized between each stage. This means that TMT only needs to focus on locations at the same time or adjacent locations to complete inference. In contrast, for the heavily occluded sequence (Fig.\ref{fig:attn_map_hd}), attention is more dispersed, and some motion tokens generate activations at various stages and ticks, indicating that these frames are regarded as ``keyframes" by the model and participate in the completion of the motion sequence.

\noindent\textbf{Additional Evaluation.} We also investigate the robustness against high-speed motion and environmental disturbances. See the \textit{supplementary materials} for more details. 

\begin{table}[t]
\centering
\caption{Ablation study of stages and the effectiveness of our Trajectory-Aware Degradation (TAD) policy. }

\begin{subtable}[t]{\linewidth}
\centering
\resizebox{\linewidth}{!}{
    \begin{tabular}{c|c|lll}
        \toprule
        \textbf{Levels $L$} & \textbf{$\mathcal{S}$} & \textbf{MPJPE} & \textbf{MPVPE} & \textbf{Accel Err} \\ \midrule
        1 & $\{32\}$ & 68.0 & 86.8 & 28.9 \\ \midrule
        2 & \multirow{2}{*}{$\{32,16\}$} & 67.3 & 86.0 & 28.8 \\
        +TAD &  & 66.9\textsubscript{\textbf{$\downarrow$0.4}} & 85.5\textsubscript{\textbf{$\downarrow$0.5}} & 28.9 \\ \midrule
        3 & \multirow{2}{*}{$\{32,16,8\}$} & 67.9 & 86.9 & 28.9 \\
        +TAD &  & \cellcolor{gray}\textbf{66.8}\textsubscript{\textbf{$\downarrow$1.1}} & \cellcolor{gray}\textbf{85.4}\textsubscript{\textbf{$\downarrow$1.5}} & \cellcolor{gray}\textbf{28.8}\textsubscript{\textbf{$\downarrow$0.1}} \\ \midrule
        4 & \multirow{2}{*}{$\{32,16,8,4\}$} & 69.0 & 88.0 & 29.0 \\
        +TAD &  & 67.3\textsubscript{\textbf{$\downarrow$1.7}} & 86.1\textsubscript{\textbf{$\downarrow$1.9}} & 28.8\textsubscript{\textbf{$\downarrow$0.2}} \\ \bottomrule
    \end{tabular}
    }
    \vspace{-0.2em}
    \caption{Effect of TAD across different numbers of stages.}
    \label{tab:ablation_stages_tad}
    \vspace{.5em}
\end{subtable}
\begin{subtable}[t]{\linewidth}
\centering
\resizebox{\linewidth}{!}{
    \begin{tabular}{c|c|ccc}
        \toprule
        \textbf{Levels $L$} & \textbf{$\mathcal{S}$} & \textbf{MPJPE} & \textbf{MPVPE} & \textbf{Accel Err} \\ \midrule
        \multirow{3}{*}{3} & $\{32,8,2\}$ & 67.6 & 86.4 & 29.0 \\
        & $\{32,16,4\}$ & 67.5 & 86.2 & 28.9 \\
        & $\{32,16,8\}$ & \cellcolor{gray}\textbf{66.8} & \cellcolor{gray}\textbf{85.4} & \cellcolor{gray}\textbf{28.8} \\ \bottomrule
    \end{tabular}
    }
    \vspace{-0.2em}
    \caption{Comparison of different stage schedules when $L{=}3$.}
    \label{tab:ablation_stage_schedules}
\end{subtable}
\vspace{-1em}
\end{table}

\subsection{Ablation Study}
\noindent\textbf{Effect of Levels and TAD.} We first investigate the influence of level settings $\mathcal{S}$ in Sec.\ref{sec:bezier_degradation} and the effectiveness of our proposed Trajectory-Aware Degradation (TAD) policy. As shown in Tab.\ref{tab:ablation_stages_tad}, incorporating TAD consistently improves performance across all numbers of levels $L$. Notably, together with Tab.\ref{tab:ablation_stage_schedules}, using $L=3$ and the schedule $\{32,16,8\}$ yields the optimal performance. This indicate that a balanced temporal resolution across stages is beneficial for capturing motion dynamics.

\noindent\textbf{Components in Progressive Motion Restoration.} To evaluate the impact of the progressive motion reconstruction stage, we conduct ablation studies on the proposed components. Our baseline is \cite{liLiDARCapLongrangeMarkerless2022}, simply replacing GRU with a Transformer to verify it is not the source of the gain. As shown in Tab.\ref{tab:ablation_components}, we first evaluate the independent contributions of the representation and the architecture. Then, by including multi-level motion tokens (m.s.), motion loss (m.l.), block-wise causal mask (b.m.), and multi-stage motion aggregator (mma.), each component brings performance gains. The combination of all components demonstrates their complementary role in improving accuracy.

\begin{table}[t]
\caption{Ablation study of the introduced components in our progressive motion reconstruction stage.}
\label{tab:ablation_components}
\resizebox{\linewidth}{!}{
\begin{tabular}{cccc|lll}
\toprule
\multicolumn{4}{c|}{\textbf{Components/Variants}} & \textbf{MPJPE} & \textbf{MPVPE} & \textbf{Accel Err} \\ \midrule
\multicolumn{1}{L{.6cm}}{Base} & \multicolumn{3}{l|}{\cite{liLiDARCapLongrangeMarkerless2022} w/ transformer}    & 79.0 & 101.0 & 42.6  \\
\multicolumn{1}{L{.6cm}}{\textit{Repr.}} & \multicolumn{3}{l|}{$\blacktriangleright$ w/ Bézier \& TAD}    & 72.3\textsubscript{\textbf{$\downarrow$6.7}} & 91.4\textsubscript{\textbf{$\downarrow$9.6}} & 30.7\textsubscript{\textbf{$\downarrow$11.9}}  \\
\multicolumn{1}{L{.6cm}}{\textit{Arch.}} & \multicolumn{3}{l|}{$\blacktriangleright$ w/ Tokens \& Mask}    & 72.2\textsubscript{\textbf{$\downarrow$6.8}} & 92.4\textsubscript{\textbf{$\downarrow$8.6}} & 30.7\textsubscript{\textbf{$\downarrow$11.9}}  \\ \midrule
m.s. & m.l. & b.m.$^*$ & mma.$^*$ & \multicolumn{3}{c}{\textcolor{Silver}{$^*$: \small\textit{Repr.} is applied for compatibility}}  \\ \midrule
$\checkmark$ &  &  &  & 78.6\textsubscript{\textbf{$\downarrow$0.4}} & 106.3 & 37.9\textsubscript{\textbf{$\downarrow$4.7}} \\
$\checkmark$ & $\checkmark$ &  &  & 76.9\textsubscript{\textbf{$\downarrow$2.1}} & 102.8 & 40.2\textsubscript{\textbf{$\downarrow$2.4}} \\
$\checkmark$ & $\checkmark$ & $\checkmark$ &  & 68.9\textsubscript{\textbf{$\downarrow$10.1}} & 87.9\textsubscript{\textbf{$\downarrow$13.1}} & 29.4\textsubscript{\textbf{$\downarrow$13.2}} \\
$\checkmark$ & $\checkmark$ &  & $\checkmark$ & 70.9\textsubscript{\textbf{$\downarrow$8.1}} & 90.1\textsubscript{\textbf{$\downarrow$10.9}} & 30.0\textsubscript{\textbf{$\downarrow$12.6}} \\
$\cellcolor{gray}\checkmark$ & $\cellcolor{gray}\checkmark$ & $\cellcolor{gray}\checkmark$ & $\cellcolor{gray}\checkmark$ & \cellcolor{gray}\textbf{66.8}\textsubscript{\textbf{$\downarrow$12.2}} & \cellcolor{gray}\textbf{85.4}\textsubscript{\textbf{$\downarrow$15.6}} & \cellcolor{gray}\textbf{28.8}\textsubscript{\textbf{$\downarrow$13.8}} \\ \bottomrule
\end{tabular}
}
\vspace{-1em}
\end{table}

\section{Conclusion}
\setlength{\parskip}{0pt}
In this paper, we presented BMLiCap, a LiDAR-based 3D human motion capture framework. The core idea is to represent and degrade human motion using Bézier curves. BMLiCap employs a Time-scale Motion Transformer to reconstruct motion across multiple temporal scales, while a Multi-level Motion Aggregator fuses these predictions and enforces temporal coherence. The proposed method is robust to sparsity, occlusion, and noise, and achieves state-of-the-art performance on four public benchmarks. We offer a new perspective on modeling human motion that can be extended to richer skeletal topologies, additional sensing modalities, and broader application scenarios.

\section{Acknowledgements}
This work was supported by the National Key Research and Development Program of China (International Collaboration Special Project, No. SQ2023YFE0102775), and National Natural Science Fund of China (Nos. U24A20330, 62361166670 and 62572242).

{
    \normalem
    \small
    \bibliographystyle{ieeenat_fullname}
    \bibliography{main}
}


\end{document}